\documentclass[sigchi]{acmart}

\usepackage{booktabs}
\usepackage{multirow}
\usepackage{graphicx}

\AtBeginDocument{%
  }

\setcopyright{acmlicensed}
\copyrightyear{2024}
\acmYear{2024}
\acmDOI{XXXXXXX.XXXXXXX}

\acmConference[Ubicomp/ISWC '24]{Make sure to enter the correct conference title from your rights confirmation email}{October 05--09, 2024}{Melbourne, Australia}
\acmISBN{978-1-4503-XXXX-X/18/06}

\begin{document}

\title[Magnitude and Rotation Invariant Detection of Transportation Modes]{Magnitude and Rotation Invariant Detection of Transportation Modes with Missing Data Modalities}

\author{Jeroen Van Der Donckt}
\authornote{Both authors contributed equally to this research.}
\orcid{1234-5678-9012}
\affiliation{%
  \institution{IDLab, Ghent University-imec}
  \city{Ghent} \country{Belgium}
}
\email{jeroen.vanderdonckt@ugent.be}

\author{Jonas Van Der Donckt}
\authornotemark[1]
\affiliation{%
  \institution{IDLab, Ghent University-imec}
  \city{Ghent} \country{Belgium}
}
\email{jonvdrdo.vanderdonckt@ugent.be}

\author{Sofie Van Hoecke}
\affiliation{%
  \institution{IDLab, Ghent University-imec}
  \city{Ghent} \country{Belgium}}
\email{sofie.vanhoecke@ugent.be}

\renewcommand{\shortauthors}{Van Der Donckt et al.}

\begin{abstract}
This work presents the solution of the Signal Sleuths team for the 2024 SHL recognition challenge. The challenge involves detecting transportation modes using shuffled, non-overlapping 5-second windows of phone movement data, with exactly one of the three available modalities (accelerometer, gyroscope, magnetometer) randomly missing.
Data analysis indicated a significant distribution shift between train and validation data, necessitating a magnitude and rotation-invariant approach. 
We utilize traditional machine learning, focusing on robust processing, feature extraction, and rotation-invariant aggregation.
An ablation study showed that relying solely on the frequently used signal magnitude vector results in the poorest performance. Conversely, our proposed rotation-invariant aggregation demonstrated substantial improvement over using rotation-aware features, while also reducing the feature vector length. Moreover, z-normalization proved crucial for creating robust spectral features.

\end{abstract}



\begin{teaserfigure}
  \includegraphics[width=\textwidth]{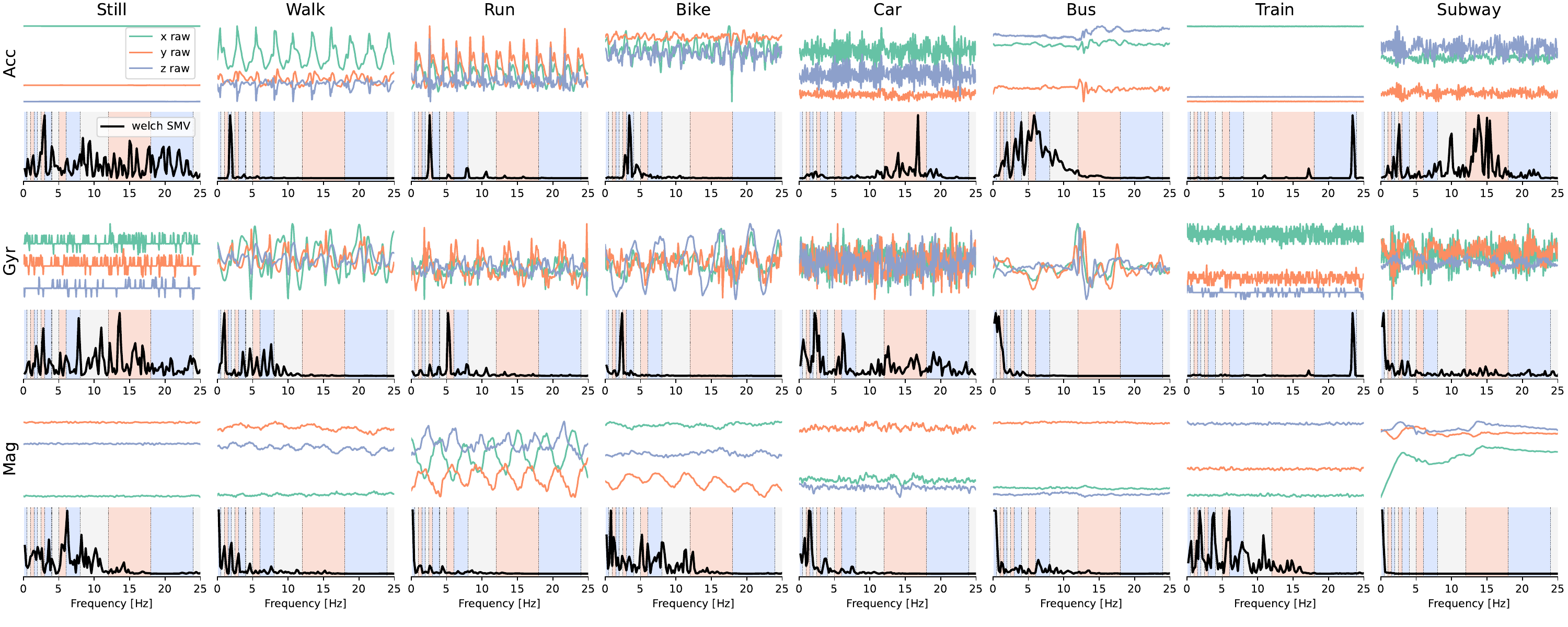}
  \vspace*{-8mm}
  \caption{Visual representation of utilized frequency bins (in the feature vector) for the power spectrum density of the signal magnitude vector per signal modality (rows) for each transportation mode (columns).}
  \Description{Each signal modality subplot displays the 3-axis signal data. Below, the Welch PSD representation is shown of the SMV aggregation of the modality. On this Welch PSD plot, the frequency bins (x-axis) are color-coded.}
  \label{fig:teaser}
\end{teaserfigure}

\received{20 June 2024}
\received[revised]{TBD}
\received[accepted]{TBD}

\maketitle

\section{Introduction}
\begin{figure*}[!t]
    \centering
    \includegraphics[width=\linewidth]{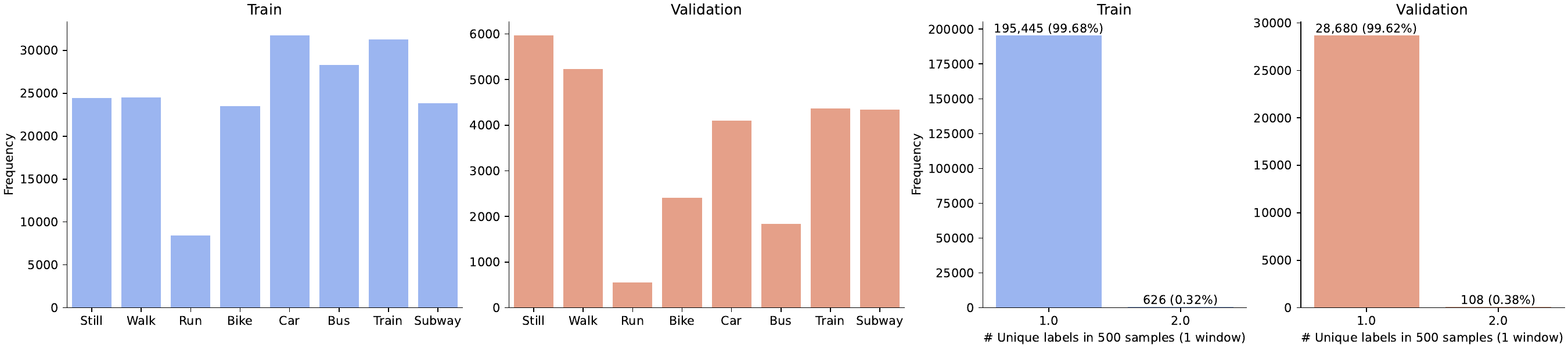}
    \vspace{-7mm}
    \caption{Label distribution for train and validation dataset (column 1-2) and number of unique labels per window (column 3-4).}
    \vspace{-3mm}
    \label{fig:label_distr}
    \Description{The first two subplots demonstrate that the train and validation datasets have nearly identical label distributions, except the "Bus" label, for which the validation set contains less data. The last two subplots showcase that there are nearly no windows (i.e., fewer than $0.4\%$ contain more than 1 label.}
\end{figure*}

Accurate and near-real-time detection of human transportation modes is crucial for applications such as automatic activity recognition on smartphones~\cite{stojchevska2024unlocking} or smartwatches~\cite{wearables_for_human_2015}, route recommendations, and context-aware service adaptations (e.g.~service switching to car mode)~\cite{mobile_phones_transp_mode}. Moreover, through ubiquitous devices such as smartphones or wearable devices, determining these transportation modes enables longitudinal behavior monitoring, which is essential for chronic disease management and follow-up~\cite{baig_systematic_2017, van2022self}.

\begin{figure*}[!t]
    \centering
    \includegraphics[width=\linewidth]{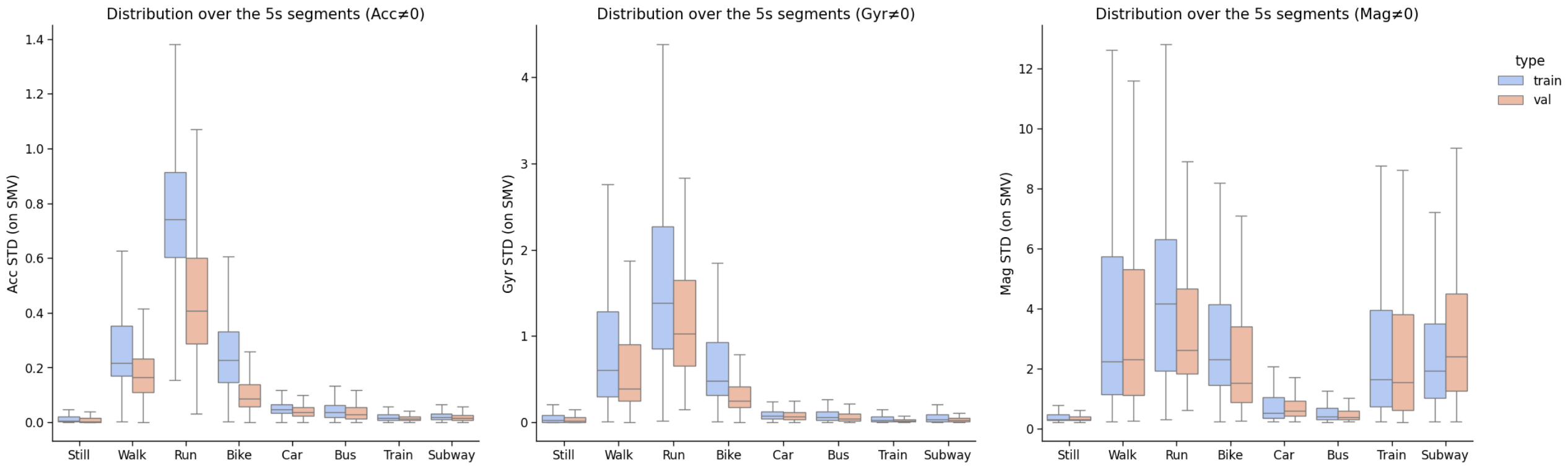}
    \vspace{-7mm}
    \caption{Data distribution (of the standard deviation - STD) for the train and validation dataset.}
    \Description{The leftmost subplot displays the distribution for the accelerometer signal across all locomotion labels via box plots. The train and validation datasets are color-coded. A higher ACC-STD is observed for the train set across most locomotion labels. The same trend is observed in the middle subplot for the gyroscope signal. The last subplot showcases this STD trend for the magnetometer signal, for which there appears to be fewer discrepancies in the distribution of the train and validation data.}
    \label{fig:train_val_std_distr}
    \vspace{-3mm}
\end{figure*}

However, creating a reference dataset that includes qualitatively annotated transportation mode data, representative of real-life settings, is challenging. 
To address this, the University of Sussex-Huawei Locomotion-Transportation (SHL) dataset was developed. The SHL dataset contains qualitatively annotated longitudinal locomotion data recorded via smartphones by three participants~\cite{wang_enabling_2019, gjoreski_university_2018}. To capture the variability of typical phone carry positions, participants carried four smartphones on different body locations (i.e., hand, torso, hip, and bag). Transportation modes in the SHL dataset were precisely post-annotated using 30-second interval images to minimize recall errors~\cite{gjoreski_university_2018}.

Given our experience with tackling multimodal time series classification using traditional machine learning, we chose to not apply deep learning for this challenge~\cite{sleep_linear_model}. Our proposed pipeline consists of (i) processing, (ii) feature extraction, and (iii) a traditional machine learning model (i.e., CatBoost - gradient boosted trees) for each missing modality. In particular, we aim to construct an efficient yet performant pipeline, allowing us to investigate the impact of various processing steps and our novel rotation-invariant feature aggregation, over different feature subsets.


\section{SHL 2024 Challenge dataset}
The 2024 SHL challenge, now in its sixth iteration, aims to detect transportation modes using multimodal sensor data from a single smartphone.
Specifically, non-overlapping 5-second windows of the smartphone's accelerometer (Acc), gyroscope (Gyr), and magnetometer (Mag) data sampled at 100Hz are provided, requiring the prediction of the current locomotion mode at a sample-level (500 predictions per window). 
This year's challenge is particularly difficult due to four key aspects.
First, the non-continuous (i.e., shuffled) test set makes post-processing predictions infeasible. This way, the organizers aim to evaluate the ability to detect locomotion modes in near-real time. 
Second, the test dataset does not provide the smartphone location (e.g., hands, hips, torso, or bag), necessitating a location-independent model.
Third, as (exactly) one of the data modalities, i.e.~Acc, Gyr, or Mag, is masked with zeros in both the validation and test datasets, the approach must be robust to missing data modalities. 
Fourth, the train dataset consists of data from a single user (subject 1), while the validation and test datasets are a mix of data from subjects 2 and 3, encouraging the development of user-independent models.

\subsection{Exploratory Data Analysis}
\begin{figure}[!t]
    \centering
    \includegraphics[width=\linewidth]{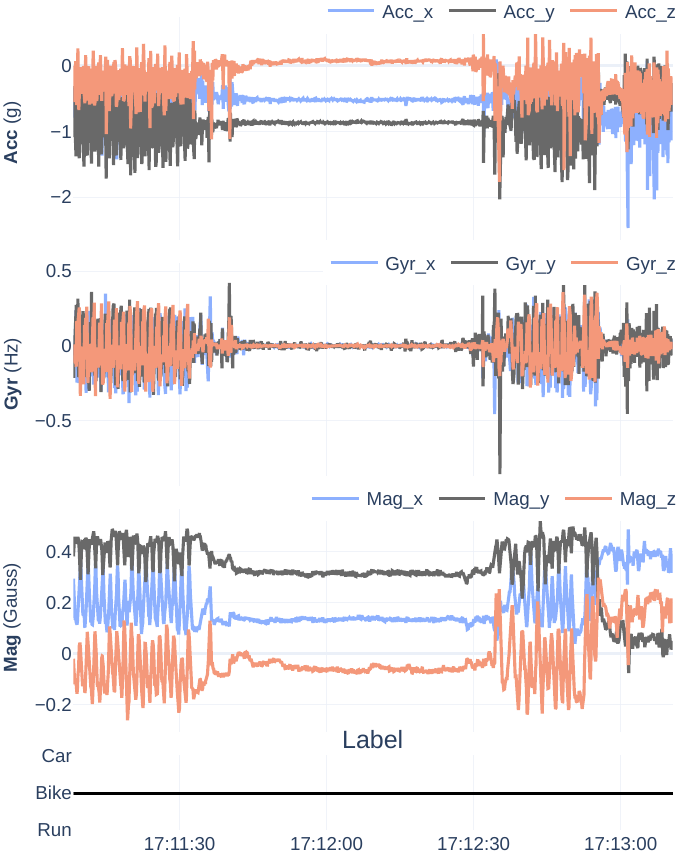}
    \vspace{-7mm}
    \caption{Exploratory time series visualization of phone movement data from train set (location = Hips) and labels.}
    \Description{Time series line chart of an excerpt of the raw train data. The lower line chart represents the label, which is "Bike" for the whole interval. In the upper magnetometer, gyroscope, and accelerometer subplots, we observe a 1-minute segment where no movement occurs. This no-movement segment likely represents the person waiting for a stoplight.}
    \label{fig:train_still}
    \vspace{-4mm}
\end{figure}

Figure~\ref{fig:label_distr} shows the distribution of transportation modes (i.e., the labels) and the number of unique modes per window in the train and validation datasets. Given that fewer than $0.4\%$ of the 5s windows contain multiple labels, we adapted the objective to predicting a single activity label for each window.

Figure~\ref{fig:train_val_std_distr} compares the distribution (standard deviation) of the signal magnitude vector (SMV) for each 5-second window across the different labels in the train and validation datasets. To ensure a fair comparison, the validation data was trimmed (per modality) to include only windows where the visualized modality was not missing. It is important to perform this comparison across different labels, as each locomotion mode can affect the feature distribution. 
From Figure~\ref{fig:train_val_std_distr}, we observe a distinct distribution shift between the train and validation data, especially for the Acc and Gyr modalities. This shift is most noticeable for the walk, run, and bike locomotion modes. In contrast, the Mag data does not demonstrate such a shift. There may be several causes for this distribution shift, such as variability between recording devices (e.g., smartphone firmware~\cite{gjoreski_university_2018}) or variability between subjects, as the train and validation datasets comprise different users. Consequently, solutions trained on solely the train data demonstrate limited generalizability. Particularly, Widhalm P. et al.~\cite{widhalm_top_2018} observed a significant drop in prediction accuracy between train and validation data. Addressing this distribution shift will therefore be a key aspect in this year's challenge.

By utilizing our \texttt{plotly-resampler} toolkit~\cite{van_der_donckt_plotly-resampler_2022}, we were capable of effectively analyzing all modalities of the train and validation data along with the labels, totaling $\pm$4.5B data points, through interactive line chart visualization. This analysis provided insights into the relationship between the training/validation data and the labels. Note that this analysis could not be performed on the test set, as this data was shuffled, thereby removing the temporal aspect. 
Figure~\ref{fig:train_still} presents an interesting excerpt from this analysis. The continuously labeled "Bike" data contains a segment of about 1 minute with nearly no movement (i.e., 17:11:30-17:12:30). This does not necessarily indicate mislabeled data as, for instance, the subject could be waiting at a stoplight. Evidently, postprocessing demonstrated substantial performance improvements (absolute 10\% gain) on the SHL dataset~\cite{wangbenchmarking2018}. 
However, relying solely on shuffled 5-second windows and being unable to use postprocessing -- which demonstrated substantial performance improvements (absolute 10\% gain) on the SHL dataset~\cite{wangbenchmarking2018} -- makes accurately classifying such small windows nearly impossible. 

\section{Algorithm pipeline}
When devising our approach, we focused on two aspects: (i) rotation invariance due to the unknown phone position in the test set, and (ii) magnitude invariance to cope with the distribution shift observed between the train and validation set (see Figure~\ref{fig:train_val_std_distr}). The resulting pipeline consists of 3 steps; (1) processing, (2) feature extraction, and (3) modeling.

\subsection{Processing}
During data loading, we scale each axis of the various modalities by fixed constants to convert them into more interpretable units. Specifically, accelerometer data is divided by 9.81 to convert m/s² to g units, gyroscope data is divided by 2$\pi$ to convert rad/s to Hz, and magnetometer data is divided by 100 to convert µT to Gauss. Figure~\ref{fig:train_still} displays these scaled signals.

Next, we aim to derive new signals (per modality) that are more robust to sensor placement and orientation. This involves computing the Signal Magnitude Vector (SMV) over the (transformed) x, y, and z axes for each modality $\mathcal{M} \in \{ Acc, Gyr, Mag \}$~\cite{wangbenchmarking2018}.
The SMV for each modality is computed using the formula:
$$SMV_{t,\mathcal{M}} = \sqrt{\mathcal{M} _{t,x}^2 + \mathcal{M} _{t,y}^2 + \mathcal{M} _{t,z}^2}$$
After computing the SMV on the raw data, we apply several axis transformations: the first-order gradient, the second-order gradient, and the integral - after which the SMV is calculated on this transformed data.

\noindent
The first and second-order gradients for each modality $\mathcal{M}$ and axis $a \in \{x, y, z\}$ are computed using the \href{https://numpy.org/doc/stable/reference/generated/numpy.gradient.html}{\texttt{numpy.gradient}} function:
$$dt^1_{t, \mathcal{M} ,a} = \frac{\mathcal{M} _{t+1,a} - \mathcal{M}_{t-1,a}}{2}$$
$$dt^2_{t, \mathcal{M} ,a} = \frac{dt^1_{t+1,\mathcal{M}, a} - dt^1_{t-1, \mathcal{M},a}}{2}$$
The integral for each modality $\mathcal{M}$ and axis $a \in \{x, y, z\}$ is computed via the \href{https://docs.scipy.org/doc/scipy/reference/generated/scipy.integrate.cumulative_trapezoid.html}{\texttt{scipy.cumtrapz}} function, based on the formula:
$$Integral_{t, \mathcal{M}, a} = \int_{0}^{t} \mathcal{M}_{\tau, a} \, d\tau $$
By computing the SMV on both the raw data and the outputs of these three axis transformations, we derive 4 rotation-invariant signals per modality: $SMV$, $SMV \left( dt^1 \right)$, $SMV \left( dt^2 \right)$, and $SMV \left( Integral \right)$. 
These four signals, together with the raw (i.e., three-axial) signals, will be considered for feature extraction in the next step.

\subsection{Feature Extraction}
\begin{table}[!tb]
    \caption{Overview of computed features per signal (N=70).}
    \vspace{-3mm}
    \label{tab:features}
\resizebox{\columnwidth}{!}{%
\begin{tabular}{@{}llll@{}}
\toprule
\multicolumn{1}{c}{\textbf{Domain}} & \multicolumn{1}{c}{\textbf{Notes}} & \multicolumn{1}{c}{\textbf{Features}} & \multicolumn{1}{c}{\textbf{n}} \\ \midrule
\multirow{6}{*}{\textbf{Spectral}} & \multirow{5}{*}{\begin{tabular}[c]{@{}l@{}}Computed on both PSD \\ and DCT magnitudes of \\ the z-normalized signals. \\ \\ So \textit{n} needs to be \\ multiplied by two. \end{tabular}} & \begin{tabular}[c]{@{}l@{}}Energy in frequency bands (Hz):\\ {[}.1-.5, .5-1, 1-1.5, 1.5-2, 2-2.5, 2.5-3, \\  3-4, 4-5, 5-6, 6-8, 8-12, 12-18, 18-24,\\   24-28, 28-32, 32-40, 40-50{]}\end{tabular} & 17 \\
 &  & Spectral centroid \& bandwidth & 2 \\
 &  & Ratio between two highest amplitudes & 1 \\
 &  & Amplitude: max, std \& skew & 3 \\
 &  & Top (peak) frequency & 1 \\
 &  & \begin{tabular}[l]{@{}l@{}}Mean, std, and skew (Hz) of top 5 \\ frequencies with highest amplitude \end{tabular} & 3 \\ \cmidrule(l){2-4} 
 & PSD & Spectral entropy & 1 \\ \midrule
\multirow{8}{*}{\textbf{Time}} & \multirow{3}{*}{Autocorrelation (ACF)} & Absolute mean, skew, std & 3 \\
 &  & Most prominent ACF frequency & 1 \\
 &  & Zero crossing rate \& slope sign change & 2 \\
 &  & Differential \& spectral entropy & 2 \\ \cmidrule(l){2-4} 
 & \multirow{5}{*}{Signal as-is} & Differential entropy & 1 \\
 &  & Mean crossing rate & 1 \\
 &  & skew \& kurtosis & 2 \\
 &  & Hjorth mobility \& complexity & 2 \\
 &  & Katz fractal dimension & 1 \\ \bottomrule
\end{tabular}%
}
\end{table}

\begin{table*}[!t]
\caption{Macro F1 scores of the different investigated feature combinations.}
\vspace{-3mm}
\label{tab:result_grid}
\begin{tabular}{l|ll|llllll|l}
\toprule
 & \multicolumn{2}{c}{\textbf{Overall}} & \multicolumn{2}{c}{\textbf{Acc = 0}} & \multicolumn{2}{c}{\textbf{Gyr = 0}} & \multicolumn{2}{c}{\textbf{Mag = 0}} &  \\
\textbf{Configuration} & \textbf{Val} & \textbf{Val$_{MV}$} & \textbf{OOF} & \textbf{Val} & \textbf{OOF} & \textbf{Val} & \textbf{OOF} & \textbf{Val} & \textbf{\# feats} \\
\midrule
rot\_inv$_{sort}$ + SMV + SMV$_{dt2}$ & 0.7244 & 0.7255 & 0.8039 & 0.7512 & 0.8538 & 0.7120 & 0.8177 & 0.6713 & 694 \\
rot\_inv$_{stat2}$ + SMV + SMV$_{dt2}$ & 0.7197 & 0.7255 & 0.8019 & 0.7474 & 0.8530 & 0.7179 & 0.8168 & 0.6720 & 556 \\
rot\_inv$_{sort}$ + SMV + SMV$_{dt1}$ & 0.7149 & 0.7195 & 0.8032 & 0.7386 & 0.8524 & 0.7105 & 0.8175 & 0.6684 & 694 \\
rot\_inv$_{stat2}$ + SMV + SMV$_{dt1}$ & 0.7146 & 0.7174 & 0.8004 & 0.7348 & 0.8523 & 0.7121 & 0.8163 & 0.6681 & 556 \\
rot\_inv$_{sort}$ + SMV & 0.7006 & 0.7048 & 0.7928 & 0.7087 & 0.8488 & 0.7052 & 0.8150 & 0.6586 & 554 \\
rot\_inv$_{stat2}$ + SMV & 0.6986 & 0.7040 & 0.7911 & 0.7045 & 0.8472 & 0.7102 & 0.8139 & 0.6562 & 416 \\
SMV + SMV$_{dt1}$ + SMV$_{dt2}$ + SMV$_{integral}$ & 0.6983 & 0.7012 & 0.7785 & 0.7335 & 0.8274 & 0.6898 & 0.7907 & 0.6356 & 560 \\
SMV + SMV$_{dt2}$ + SMV$_{integral}$ & 0.6940 & 0.6969 & 0.7734 & 0.7298 & 0.8231 & 0.6804 & 0.7904 & 0.6349 & 420 \\
SMV + SMV$_{dt1}$ + SMV$_{integral}$ & 0.6903 & 0.6919 & 0.7723 & 0.7193 & 0.8226 & 0.6877 & 0.7888 & 0.6296 & 420 \\
SMV + SMV$_{dt1}$ + SMV$_{dt2}$ & 0.6889 & 0.6952 & 0.7654 & 0.7240 & 0.8292 & 0.6842 & 0.7860 & 0.6392 & 420 \\
SMV + SMV$_{dt2}$ & 0.6858 & 0.6876 & 0.7586 & 0.7190 & 0.8239 & 0.6694 & 0.7857 & 0.6339 & 280 \\
SMV + SMV$_{dt1}$ & 0.6832 & 0.6864 & 0.7566 & 0.7073 & 0.8258 & 0.6822 & 0.7860 & 0.6305 & 280 \\
rot\_inv$_{sort}$ & 0.682 & 0.6828 & 0.7840 & 0.7031 & 0.8226 & 0.6728 & 0.7924 & 0.6312 & 414 \\
rot\_inv$_{stat3}$ & 0.6751 & 0.6749 & 0.7790 & 0.6893 & 0.8181 & 0.6725 & 0.7892 & 0.6235 & 414 \\
rot\_inv$_{stat2}$ & 0.6744 & 0.6758 & 0.7795 & 0.6913 & 0.8186 & 0.6716 & 0.7890 & 0.6221 & 276 \\
raw & 0.6681 & 0.6695 & 0.7828 & 0.6782 & 0.8235 & 0.6700 & 0.7937 & 0.6134 & 420 \\
SMV + SMV$_{integral}$ & 0.6579 & 0.6615 & 0.7450 & 0.6680 & 0.8038 & 0.6632 & 0.7806 & 0.6131 & 280 \\
SMV & 0.6511 & 0.6542 & 0.7215 & 0.6564 & 0.8036 & 0.6547 & 0.7728 & 0.6166 & 140 \\
\bottomrule
\end{tabular}

\end{table*}

Table~\ref{tab:features} provides an overview of the extracted features for each available signal. Our \texttt{tsflex} toolkit was used for convenient feature extraction~\cite{van_der_donckt_tsflex_2022}. 
Entropy, fractal, and Hjorth features were computed using the \texttt{antropy} toolkit~\cite{vallat_antropy}, while other features were derived from functions provided by the \texttt{numpy} and \texttt{scipy} libraries~\cite{harris_array_2020, virtanen2020scipy}. 

Most spectral features represent frequency-band energy. The frequency bounds of these bands were determined using a visual analytic approach, as illustrated in the teaser Figure~\ref{fig:teaser}.

To create a more magnitude-invariant feature set, we refrained from including magnitude-related features from the time domain (e.g., quantiles, peak-to-peak, mean, std). Moreover, we apply z-normalization before computing the Discrete Cosine Transform (DCT) and Power Spectral Density (PSD) features. We hypothesize that z-normalization can make the spectral representation features less sensitive to amplitude variations, enhancing magnitude robustness and thereby reducing susceptibility to distribution shifts.

In total, 70 features were extracted per signal, resulting in 980 features overall: 2 modalities (as one of the three is always missing) $\times$ (3 raw + 4 processed signals) $\times$ 70 features. 

\subsection{Model}
We focused on traditional machine learning models, specifically investigating the performance of CatBoost, a gradient-boosted trees algorithm known for its strong performance without parameter tuning~\cite{prokhorenkova2018catboost}, making it suitable for an ablation study on feature subsets. We limited the CatBoost model to 1,000 iterations (trees) and used "Balanced" as the \texttt{auto\_class\_weight} parameter. To cope with missing data, we created a new model for each missing modality configuration by excluding the missing modality from the (training) feature vector. Consequently, three different models were constructed, each using only two modalities. 

\subsection{Postprocessing}\label{sec:postprocess}
In contrast to most traditional approaches, we did not train on the entire train dataset to make final predictions on the test data. Instead, we leveraged K-fold cross-validation (CV), making predictions after fitting each fold, and then aggregated these predictions by utilizing majority voting (MV). We deliberately opted for a 3-fold CV, as using an odd number as K results in only 50\% data overlap across the training folds. Using this approach, we effectively perform bootstrapping through CV on the training data. 

\subsection{Rotation Invariant Aggregation}
\begin{figure*}[!t]
    \centering
    \includegraphics[width=\linewidth]{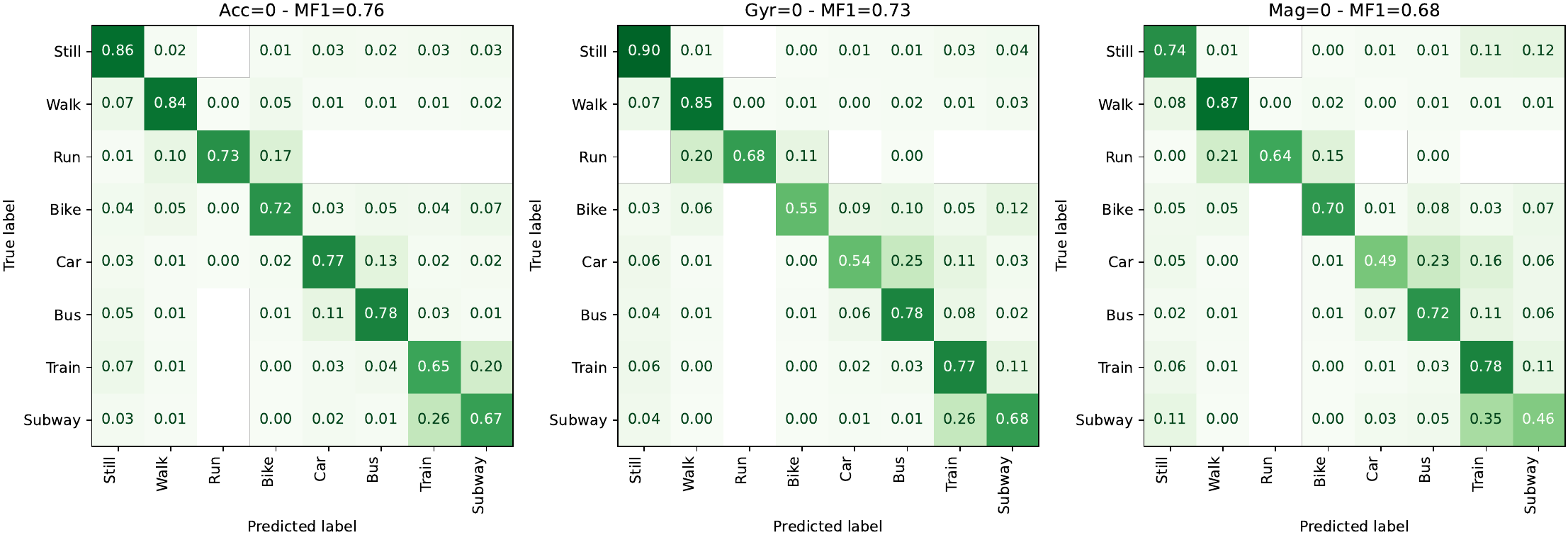}
    \vspace{-7mm}
    \caption{Confusion matrix for the final model (i.e., rot\_inv$_{stat2}$ + SMV + SMV$_{dt2}$) on validation dataset (including all 4 phone locations). The validation predictions were obtained using the 3-fold MV method, as described in Section~\ref{sec:postprocess}.}
    \label{fig:cm_3_cols}
    \vspace{-3mm}
\end{figure*}

To improve the robustness of the feature set computed on the raw axis signals (i.e., non-SMV-transformed), we proposed and examined three novel \textit{rotation-invariant statistical aggregation} approaches. Specifically, the raw $\{x, y, z\}$ signal features are condensed to summary statistics using: \texttt{stat2}: $\{$ mean, std $\}$, \texttt{stat3}: $\{$ mean, std, skew $\}$, or \texttt{sort}: $\{$ min, mid, max $\}$. These approaches effectively discard axial information while retaining the overall feature information. Note that \texttt{stat2} converts the three $\{x, y, z\}$ features into two summary statistics (resulting in a 1/3 compression ratio), while \texttt{stat3} and \texttt{sort} maintain the same input-output ratio.

Sensor orientation affects the sign of axial measurements. For instance, flipping a sensor along one axis produces opposite signs for the measurements along that axis. Consequently, rotation-invariant statistical aggregation is only relevant for sign-invariant features (e.g., kurtosis, mean crossing rate, and spectral domain features). As such, from Table~\ref{tab:features}, skewness is the only feature that is influenced by the sign, resulting in it being discarded when performing the rotation-invariant aggregation.


\section{Experimental results and discussion}\label{sec:exp_results}
Experiments were performed on a server computer (Arch Linux), with an AMD Ryzen 5 2600x CPU, 48GB of DDR4 RAM, and an Nvidia RTX 2070 GPU.
Table~\ref{tab:result_grid} presents the out-of-fold (OOF) and validation (Val) macro F1 scores for the various investigated feature combinations. Each experiment (represented by a row in the table) involved training 12 models: 4 (3 folds + 1 full fit) x 3 (for the 3 missing modalities), taking a total of $\pm$10 minutes to complete. Inference on the validation set took $\pm$2 seconds, including 3-fold MV, with processing and feature extraction taking $\pm$ 2 minutes.

A first notable observation is that using only the \texttt{SMV} signal results in the worst validation performance. Yet, \texttt{SMV} has been commonly employed as the sole processing configuration in many studies~\cite{wangbenchmarking2018, widhalm_top_2018}.
Second, our proposed rotation-invariant statistical aggregation approaches (i.e. \texttt{rot\_inv$_{stat2}$}, \texttt{rot\_inv$_{stat3}$}, and \texttt{rot\_inv$_{sort}$}) all outperform the \texttt{raw} configuration (i.e., no feature aggregation) and the \texttt{SMV}) processing. These findings highlight the potential of statistical aggregation to effectively remove the axial (i.e., rotation-aware) information while preserving overall feature information. 
Third, we observe a consistent performance improvement when applying cross-fold-based majority voting ($Val_{MV}$) compared to a model fully trained on the train set (Val).

Analyzing the performance of SMV computed on gradient and integral axis transformations reveals that incorporating the second-order gradient ($dt2$) yields the highest performance boost, followed by the first-order gradient ($dt1$), with the $integral$ contributing the least. Interestingly, combining $dt1$ and $dt2$ provides a lower boost than combining either gradient transformation with the $integral$. This might be attributed to the high correlation between the gradient transformations, making their combination less informative.

When combining the rotation-invariant statistical aggregation with the SMV and the SMVs on the gradients/integral axis transformations, we observe similar trends. Among the rotation-invariant statistical aggregations, \texttt{stat2} and \texttt{sort} exhibit the best performance. Whereas again, \texttt{dt2} performs slightly better than \texttt{dt1}, with \texttt{integral} resulting in the smallest improvement.

Although not presented in Table~\ref{tab:result_grid}, we also investigated the impact of z-normalization before computing the PSD and DCT features. On average, using z-normalization resulted in a consistent 1-1.5\% (absolute) improvement in macro F1 on the validation data.




\subsection{Impact Distribution Shift}
A consistent trend observed in Table~\ref{tab:result_grid} and in Figure~\ref{fig:cm_3_cols} is that missing magnetometer data (i.e., \texttt{MAG = 0}) results in the poorest validation performance, 
despite having a comparable OOF score to the two other scenarios. This may be attributable to the magnetometer being the only modality that does not suffer from a substantial distribution shift, as shown in Figure~\ref{fig:train_val_std_distr}. 
The smallest drop in performance between the OOF and the Val scores is observed for the \texttt{ACC = 0} case, likely due to the accelerometer data exhibiting the largest distribution shifts, as indicated by Figure~\ref{fig:train_val_std_distr}.

\subsection{Final Model}
For the final model, we selected the rot\_inv$_{stat2}$ + SMV + SMV$_{dt2}$ feature configuration, as the rot\_inv$_{sort}$ variant of this configuration resulted in the same Val$_{MV}$ score while having a larger feature vector (see Table~\ref{tab:result_grid}).

Figure~\ref{fig:cm_3_cols} shows the confusion matrices for this final model. An unexpected observation is that the models do not distinguish "Run" that well. Given the pronounced frequency-component of running, as illustrated in Figure~\ref{fig:teaser}, we anticipated better performance in recognizing this activity. We hypothesize that training on data from only one user might have caused the model to capture that user's specific running behavior too closely, as substantiated by the large distribution shift for running, observed in Figure~\ref{fig:train_val_std_distr}. Consistent with previous research, we observe typical train-subway and car-bus confusion~\cite{wangbenchmarking2018}. Finally, the availability of gyroscope data appears to enhance the predictive performance for the "Bike" class.



\subsection{Improving the Test Score}
To generate the final test predictions, we retrained our final models (3-fold CV) using several tricks to improve the test score over the above-reported validation score:
\begin{itemize}
    \item Exclude hand location: According to the challenge description, the test dataset comprises only torso, hips, and bag locations. Analysis of prediction errors per location indicated that hand location performed the worst, aligning with the observation of Gjoreski et al.~\cite{gjoreski_university_2018}. Experimental results indicate an improvement in Val$_{MV}$ from 0.7255 to 0.7506.
    \item Train on validation data: Since there were no restrictions on what data may be used for the training, including the validation during training should allow the model to better capture (any) remaining distribution shift and thereby learn the variability across multiple subjects.
\end{itemize}


\section{Conclusion}
This work presents the findings and final approach of the "Signal Sleuths" team for the 2024 Sussex-Huawei Locomotion-Transportation recognition challenge. During exploratory data analysis, we identified a substantial distribution shift between train and validation/test data, making this a key challenge in this year's edition. 
To address this, we utilized a traditional machine learning pipeline, enabling us to perform an ablation study on various magnitude and rotation-invariant feature configurations. In particular, we performed an ablation study on (i) including features from different processed signals as well as (ii) a novel approach in which we make rotation-aware signals rotation-invariant through statistical aggregation, and (iii) performing z-normalization prior to spectral feature computation.
Results indicated that solely relying on SMV yields the poorest performance.
Moreover, our proposed rotation-invariant statistical aggregation demonstrated a substantial improvement over using rotation-aware features or SMV alone, with the added benefit of reducing the feature vector length, underscoring the efficacy of this novel technique. 
Furthermore, z-normalization of the data proved to be beneficial when creating robust spectral features.
Our final model consists of the best feature subset combination, retrained on both the training and validation data to better capture inter-user variability, while also excluding the hand data as this location is not present in the test data. 
The recognition result for the testing dataset will be presented in the summary paper of the challenge~\cite{wang_2024_shl_summary}.

\begin{acks}
Jonas Van Der Donckt (1S56322N) is funded by a doctoral fellowship of the Research Foundation Flanders (FWO).
\end{acks}

\bibliographystyle{ACM-Reference-Format}
\bibliography{references}



\end{document}